\documentclass[10pt,twocolumn,letterpaper]{article}

\usepackage{cvpr}              

\usepackage{graphicx}
\usepackage{amsmath}
\usepackage{amssymb}
\usepackage{booktabs}
\usepackage{algorithm}
\usepackage{algpseudocode}
\usepackage{placeins}
\usepackage{comment}

\usepackage[pagebackref,breaklinks,colorlinks]{hyperref}

\usepackage[capitalize]{cleveref}
\crefname{section}{Sec.}{Secs.}
\Crefname{section}{Section}{Sections}
\Crefname{table}{Table}{Tables}
\crefname{table}{Tab.}{Tabs.}

\def\cvprPaperID{*****} 
\def\confName{CVPR}
\def\confYear{2023}

\begin{document}

\title{Facial asymmetry: A Computer Vision based behaviometric index for assessment during a face-to-face interview}

\author{
Shuvam Keshari\\
{\tt\small skeshari@utexas.edu}
\and
Tanusree Dutta\\
{\tt\small tanusree@iimranchi.ac.in}
\and
Raju Mullick\\
{\tt\small raju.mullick@gmail.com}
\and
Ashish Rathor\\
{\tt\small ashishrathor72@gmail.com}
\and
Priyadarshi Patnaik\\
{\tt\small priyadarshi1@yahoo.com}
}
\maketitle

\begin{abstract}
Choosing the right person for the right job makes the personnel interview process a cognitively demanding task. Psychometric tests, followed by an interview, have often been used to aid the process although such mechanisms have their limitations. While psychometric tests suffer from faking or social desirability of responses, the interview process depends on the way the responses are analyzed by the interviewers. We propose the use of behaviometry as an assistive tool to facilitate an objective assessment of the interviewee without increasing the cognitive load of the interviewer. Behaviometry is a relatively little explored field of study in the selection process, that utilizes inimitable behavioral characteristics like facial expressions, vocalization patterns, pupillary reactions, proximal behavior, body language, etc. The method analyzes thin slices of behavior and provides unbiased information about the interviewee. The current study proposes the methodology behind this tool to capture facial expressions, in terms of facial asymmetry and micro-expressions.  Hemi-facial composites using a structural similarity index were used to develop a progressive time graph of facial asymmetry, as a test case. A frame-by-frame analysis was performed on three YouTube video samples, where Structural similarity index (SSID) scores of 75\% and more showed behavioral congruence. The research utilizes open-source computer vision algorithms and libraries (python-opencv and dlib) to formulate the procedure for analysis of the facial asymmetry.

\end{abstract}

\section{Introduction}
\label{sec:introduction}
The popularity of face-to-face interviews for personnel selection cannot be denied. An employment interview is often the first contact between the interviewee (seeking employment) and the interviewer (employer). Many applicants recognize the importance of this initial contact and make efforts to ensure a positive outcome. The nonverbal behaviors that an applicant exhibits during this stage of the interview are often a major determinant of the kind of impression that is made. Research suggests that facial emotion expression does serve as communicative signals and as elicitors, facial emotions also influence judgments and preferences of the perceiver \cite{ruys2008secret}. These facial expressions at times influence the employer's decision to hire. Between the interviewee’s verbal and nonverbal behavior, it is the latter that can provide personal information without the conscious awareness of the respondent. In fact, emotional expressions are often a reflection of the internal mental state \cite{lee2014optical}. Such information is frequently sought by interviewers, but at the same time, identifying discrepancies between content (content) and intent (nonverbal) of the interviewee’s speech or facial expressions during the ongoing personnel selection interview itself is cognitively overbearing. This compromises the objectivity of assessment in a selection interview process, making the assessment vulnerable to subjectivity, and behavioral and cognitive bias. 
The face of an individual is a canvas on which an individual expresses and at times hides certain emotions. However, according to Paul Ekman (Frank, 2007 c.f \cite{porter2008reading}), some facial communication cannot be controlled and they give away true information. This can be detected by a trained observer only.  This proves to be a challenge in any interaction as every observant is not trained at all times. To address this concern, we propose the development of behaviometric tool based on machine learning to assist interviewers in making objective decisions about the interviewee. We present the details of this tool along with the results of a pilot study conducted with the help of this tool, to qualify the effectiveness of the tool.

\subsection{Behaviometry}
Presently psychometry is popularly used to understand human behavior. Psychometry involves profiling of behavior based on the measurement of cognitive capabilities, affective predisposition, behavior orientation including personality. Alongside psychometry, biometric data is also used at times to understand human behavior. While biometry data fails to reflect the mental or emotional states of an individual, psychometric data fail to reveal objective verifiable real-time human behavior. Although psychometric tests are developed based on behavior samples and are internally validated, items do not always match a visual image or reflect a signature behavior for assessment which is good for external validity. The science of behaviometry, which uses thin slices of nonverbal behavior as the measure of mood states, is fast evolving and appears to be a probable solution to the existing problem.
Behaviometry interfaces between psychometrics and biometry by utilizing unique behavioral signatures like the facial expression of emotion, ocular motion, gait pattern and body language, vocalization, proximal behavior (use of physical space during communication), and other forms of nonverbal behaviors. This study limits the use of behaviometry to analyze facial expressions to assist objective understanding of human behavior during personnel selection. The term behaviometry is made up of two words: behavior and metric. While behavior relates to the unique identification of an individual based on behavior characteristics, metric indicates measurement. Behaviometry is an objective measurement of the unique and common behavior pattern exhibited by an individual. It helps in reinforcing, faking, substituting, or at times contradicting verbal behavior.
The reason for developing behaviometry as a tool to capture and analyze Facial Emotion during an interview situation is because for many years behaviometry has stood the test of time to understand psychopathology. Its application to understanding normal human behavior however has been restricted.

\section{Related Work}
\label{sec:micro}
\subsection{Microexpressions of emotions and facial asymmetry }

The face is a salient and powerful representation of an affective state \cite{matsumoto2001culture}. It helps portray the importance of emotions essential for basic survival, be it physical, psychological, or social and communication \cite{mandal2004laterality}. The face is exposed to others for the facilitation of social system interaction. The amount (especially for a brief moment) and kind (emotional, attitudinal) of information that it conveys is easy to comprehend. \cite{ekman1975unmasking} noted ‘the face is a multi-signal as well as multi-message system and one of the most important messages that it renders is emotion’. It is assumed that expressions of the face have enhanced innate disposition to reflect true psychic reality through genuinely felt emotional expressions. With the advancement of technology involving computer vision and artificial intelligence, it has become possible today to detect facial behaviors. Furthermore, brief expressions in dynamic conditions such as during interviewing that easily evade our notice can also be captured using computer vision.
A second reason for the paucity of attempts to quantify facial behavior is the sheer volume of expressions that mimetic muscles of the face can elicit. These muscles are activated by facial nerves, which cause contractions resulting in observable movements. Visible facial actions are blocks of skin motions or wrinkles, the configuration of which changes with the change in mood state. Since mood state is rarely reflected in pure form, the face reveals compound emotion (blended with other emotions like happiness, sadness, fear, anger, surprise, or disgust, (see \cite{du2014compound}), the nature and variety of which depends on cultural display rules (like intensification, de-intensification, neutralization, masking (see \cite{matsumoto2001culture}), personality disposition, clinical state, demographic characteristics (like gender, ethnicity, age), and other factors. Thus, it is challenging to calibrate each expression for objective assessment.  
Unlike Psychometry which is a behavioral assessment of the subject by asking relevant questions, behaviometry is the analysis of thin slices of human behavior by recording nonverbal cues of micro-expression of the face or body and analyzing them with the help of machine learning.

\subsection{Lateralization of facial expression: Quantifying facial asymmetry}
Quantifying facial asymmetry is a challenging task. Available measures are either scanty or lack the exactitude that would help measure subtle changes across the face. No doubt,  facial expression recognition systems have tried to address these issues, but they perform well when the intensity of expressions is high \cite{pantic2000expert}.
 
It was Darwin’s \cite{darwin1890expression} classical work that suggested that the two halves of the face differ in terms of facial expression. Wolff \cite{wolff1933experimental} further suggested that the right side is the social face that expresses desirable emotions unlike the left side of the face that expresses personal emotions. The left side of the face is governed by the right hemisphere, which specializes in processing emotions.
 
Asymmetry is a structural description that cannot be captured by a local measure. We propose the development of a split-face composite method. It is a process wherein a full face is reconstructed by combining the lateral half of one side of the face with its mirror opposite \cite{mandal2000side}. The right hemisphere of the brain, which exerts contralateral motor control on the left side of the face, is also considered dominant in the processing of emotion \cite{borod1997neuropsychological}. Empirical evidence suggests that the left side of the face is more affect-laden as opposed to its counterpart (see \cite{asthana1997hemiregional, sackeim1978emotions}). The right side of the face is more under the voluntary control of the expresser and displays socially desirable emotional expressions in healthy adults (for a detailed review on the subject, see \cite{murray2015asymmetry}). Compelling evidence for this proposition has come from a series of experiments conducted by this author on normal subjects \cite{mandal1995asymmetry} and brain-damaged patients \cite{mandal1999hemiregional}.

\section{Methodology}
\label{sec:methods}

\subsection{Hypothesis}
Based on the available literature, we hypothesize that a composite of the left-left (L-L) side of the face developed using the split-face composite method would be indicative of personal emotions while a composite of the right-right (R-R) face would be indicative of social expression. A discrepancy in the expression of an L-L and R-R facial composite would suggest faking while synchrony or enhancement of L-L and R-R facial composite would suggest the authenticity of emotion. By using a computer-based program that utilizes open-source codes, we have demonstrated an example of how we can generate the L-L and R-R facial composites dynamically from a video, analyze how similar they are, and predict whether the emotion expressed is genuine.
 
The purpose of the present study is to demonstrate the utility of examining the facial asymmetry of the candidate during the interview, develop an artificial intelligence-based program to decipher the mental state or emotion and utilize the hemifacial asymmetry index to supplement the judgment of the interviewer.


\subsection{Sample}
The sample of the study consisted of three video snippets available on the internet as open source.  These videos were structured interview sessions and chosen randomly from a set of such sessions. The face of the interviewee was discernible in each session but they had slightly different placement of the camera, resulting in videos where the interviewee was either directly facing the camera or slightly tilted away from the camera. Theoretically, a video is a combination of continuous image frames put together in motion. To detect the presence of a face in the video clips, and henceforth its asymmetry, each video clip was run through the software (available here on Github: \cite{keshariS/FacialAsymmetry}), and analysis was done frame by frame using algorithms that operate only on a single image. 

\subsection{Algorithm flow}
\label{algo}
The sequence of activities that followed the selection of the 3 videos is outlined below:
\begin{enumerate}

    \item We begin by extracting individual frames from the videos. The frame rate of the videos was 60fps, meaning every second there are 60 different frames to analyze, which results in every frame being about 17 milliseconds long.
    
    \item Once the frames have been extracted, the `dlib' \cite{dlib_pypi} face detector (which is an open-source library) was used to detect faces from each frame in a video file. There could be occasional errors where the program fails to detect the presence of a face in a specific frame.
    
    \item For those frames where a face was successfully detected, `dlib' identified 68 facial landmark points which represent the exact location and orientation of the facial features (as shown in Figure ~\ref{fig:land68}). The added advantage of using `dlib' is that it also takes care of the alignment of an image (a slightly tilted image). This is one of the primary reasons for choosing `dlib' from other state-of-the-art face detector algorithms such as Viola-Jones.

\FloatBarrier
\begin{figure}[h]
\centering
\includegraphics[width=0.4\textwidth]{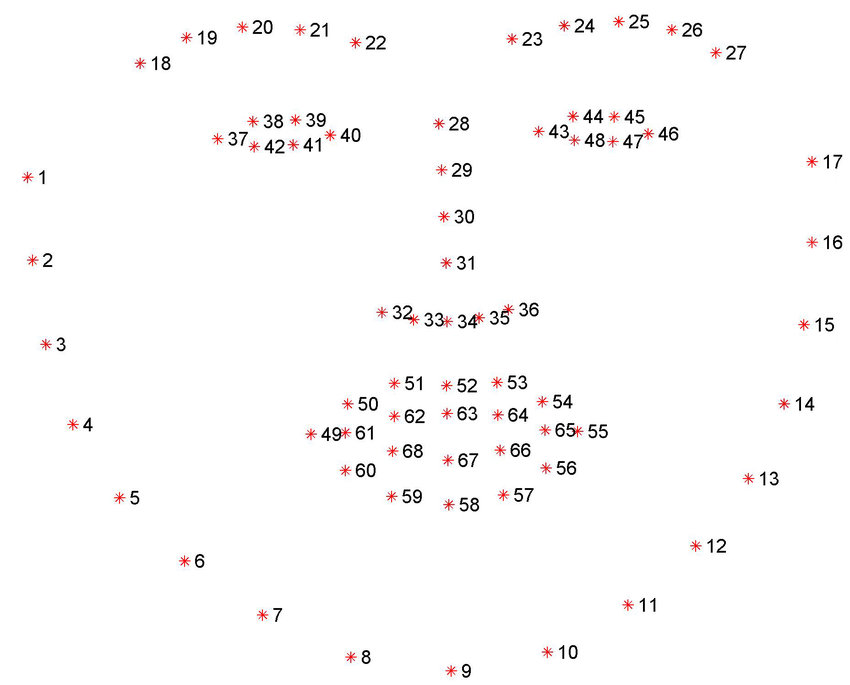}
\caption{68 facial landmarks defined in dlib}
\label{fig:land68}
\end{figure}
\FloatBarrier

    \item The algorithm then extracts pixels only from the relevant face region from the frame. It then measures the left side of the face and then makes a copy of all pixel values from the left side of the face and flips it to the right side to create an artificial Left-Left (L-L) composition of the face. The same is followed for creating the right-right (R-R) composition of the face. 
    
    \item Then the structural similarity index (SSID) technique was used to verify the similarity between the two artificially constructed faces i.e. the L-L and R-R faces from each frame. This SSID technique returns a value between 0 and 1. 1 or close to 1 indicates that the faces are more similar whereas 0 or close to 0 indicates dissimilarity between faces.
    
    \item As previously mentioned, there can be occasional errors when no face is detected in a frame (possible reasons could be that the subject was looking away from the camera momentarily). Our algorithm takes care of these image frames where no faces were detected using SSID: it returns a negative value (-0.1) (where in reality either a greatly tilted or rotated face was present in the frame or the camera focused away from the subject's face).
    
    \item The steps above are done for all the frames in the video and the corresponding SSIDs are recorded. Hence, the methodology generates a visual graph (see Figure ~\ref{fig:res}) after analyzing each video based on SSID (represented as a percentage on the y-axis) and time (duration of the clip, also normalized to a percentage on the x-axis).

    \item This graph in a way represents instances where the L-L and R-R face composites were highly dissimilar. To make sense of the generated graph, we first need to ascertain the baseline asymmetry value (an SSID) for the subject being interviewed. This is because every individual will have some degree of inherent/irreducible dissimilarity in the face composites and this dissimilarity varies from person to person. This baseline SSID can be determined from a static image of the subject during a neutral emotional state (when the subject is calm/neutral outside of the interview setting). We didn't have access to this for the videos we studied, however, we assumed the baseline would lie somewhere between the highest and lowest peaks of the generated graph.
    
    \item The algorithm then compares this baseline asymmetry value with the graph for the entire video. Those time instances where the SSID is significantly less than the baseline (the lowest peaks) indicate a high degree of facial asymmetry which further suggests incongruity in emotional expression. 

\end{enumerate}
To summarize, by using Python (open-source code libraries), the composite of L-L and R-R faces were developed from the recorded video file to analyze the possible incongruencies across the time domain.

\FloatBarrier
\begin{figure}[h]
\centering
\includegraphics[width=0.47\textwidth]{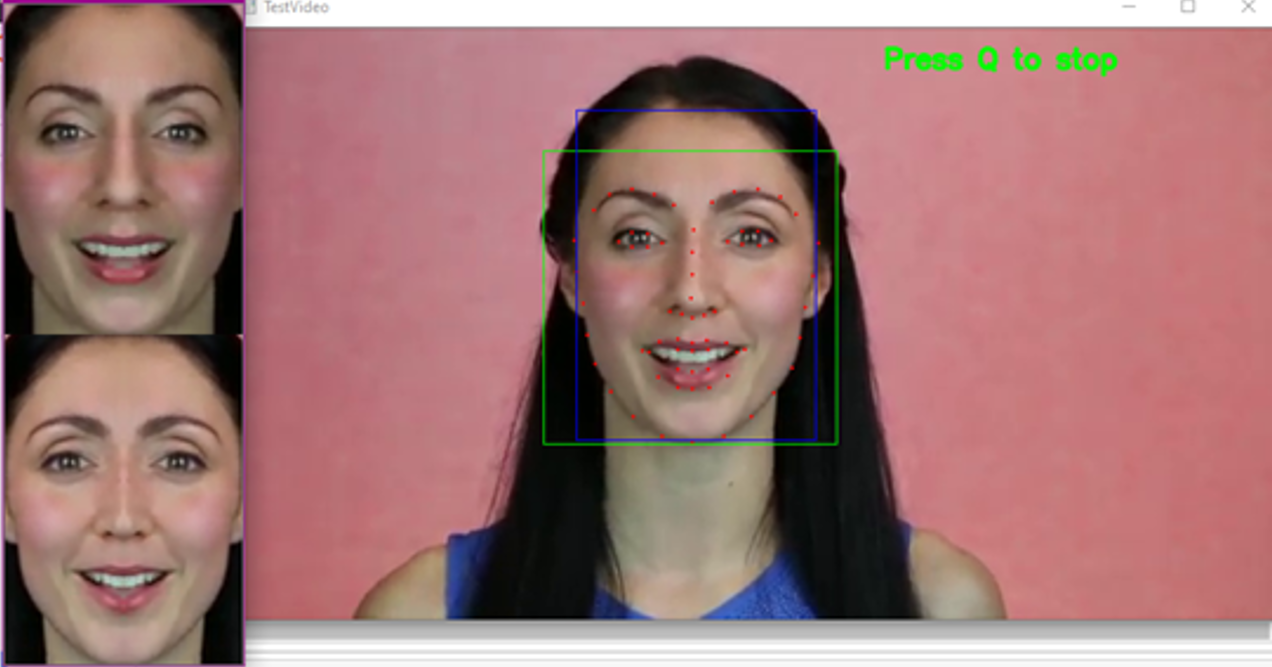}
\caption{A sample frame of video snippet 1, where both the L-L (upper left corner) and R-R (lower left corner) along with the original (middle of the window) faces are being displayed on the image (ref: \cite{video1})}
\label{fig:f1}
\end{figure}
\FloatBarrier

\FloatBarrier
\begin{figure}[h]
\centering
\includegraphics[width=0.47\textwidth]{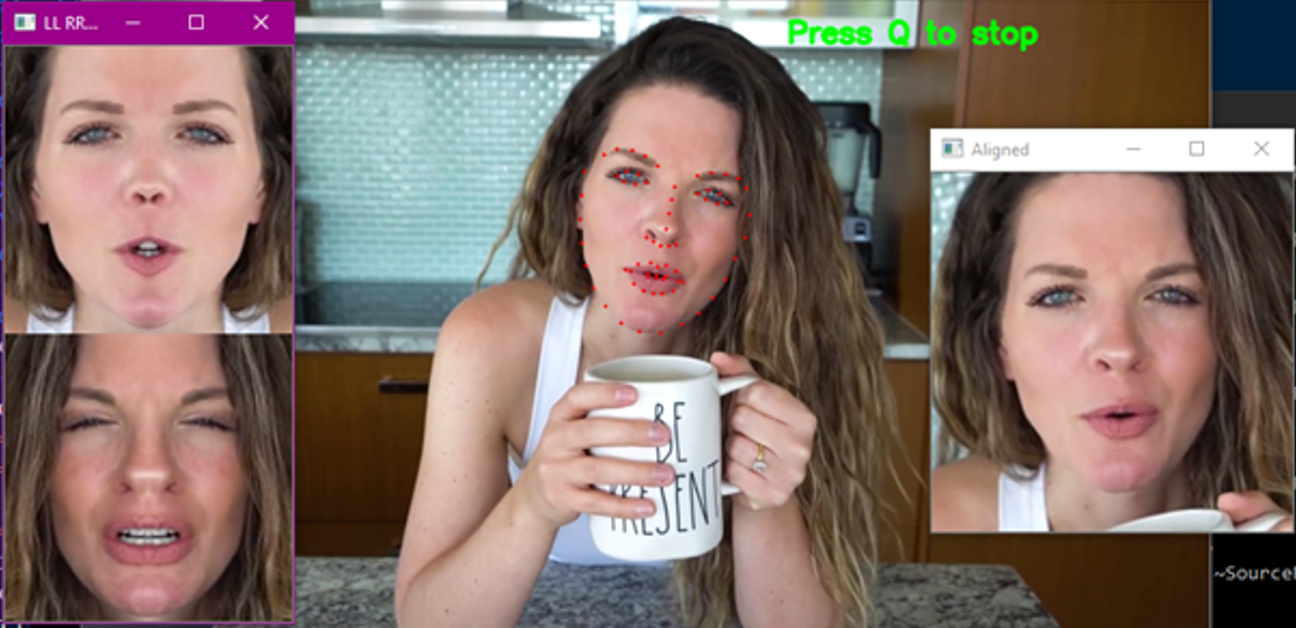}
\caption{A sample frame of video snippet 2 that demonstrates the implementation of aligning the face first and then constructing the L-L (upper left corner), R-R (lower left corner) along with the original (middle of the window), and aligned (extreme right) faces.}
\label{fig:f2}
\end{figure}
\FloatBarrier

\FloatBarrier
\begin{figure}[h]
\centering
\includegraphics[width=0.47\textwidth]{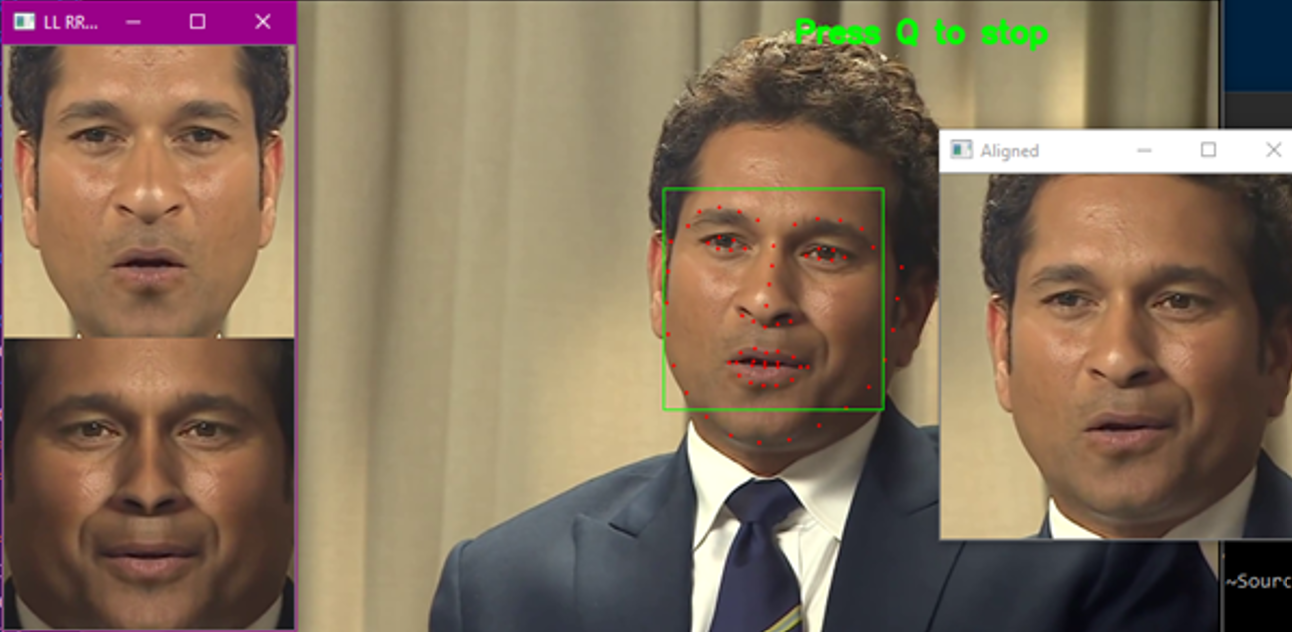}
\caption{A sample frame of video snippet 3 that demonstrates the highly tilted face:  generated L-L (upper left corner), R-R(lower left corner) along with the original(middle of the window), and aligned (extreme right) faces.}
\label{fig:f3}
\end{figure}
\FloatBarrier

\section{Results}
\label{sec:results}
\begin{figure}[h]
\centering
\includegraphics[width=0.47\textwidth]{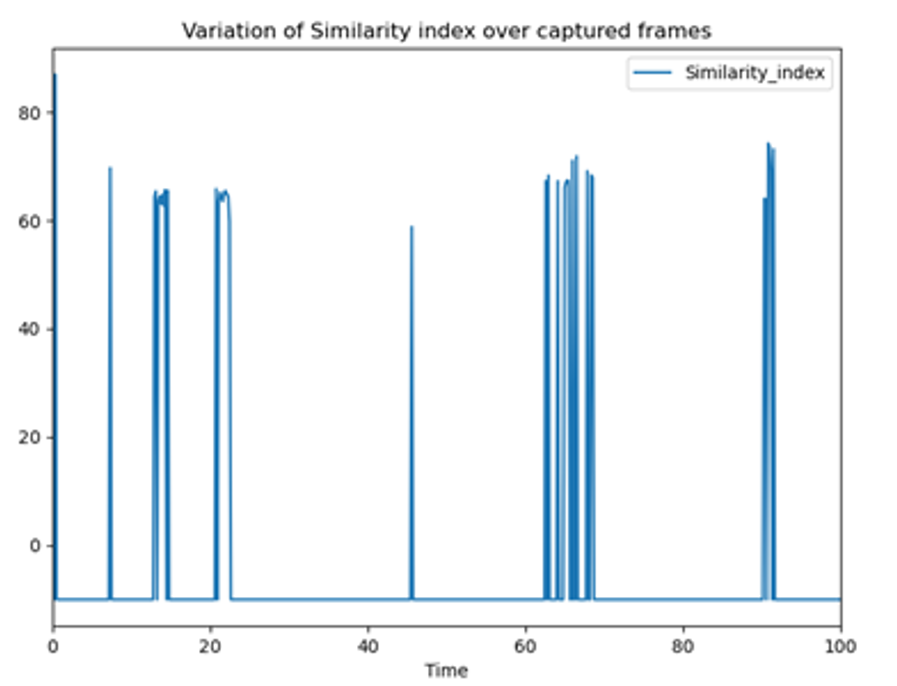}
\caption{Generated graph for a small duration of a video clip. The X-axis represents time (clip duration) and the y-axis shows the SSIDs in each frame (both values converted to percentage). Values of SSID less than 0 means that the face was either not detected or was not well aligned}
\label{fig:res}
\end{figure}

Several results were observed from the methodology that was followed. Although similar methods were followed for all 3 videos, we discuss one of the video files here:

\begin{enumerate}

    \item After analyzing one of the video files as per section \ref{algo}, we can see in Figure ~\ref{fig:res} that around the 45\% time duration of the clip, we have the lowest peak. This could indicate that the subject was trying to hide genuine emotions during that time. This assumes that the SSID value for the `baseline asymmetry' between the L-L and R-R faces lies between the highest and lowest peaks.
    
    \item A sudden fall of the SSID may represent an incongruity of the subject at that time and may be used as complementary evidence. The questions that were being asked during that time should be taken into consideration to arrive at conclusions.
Results also suggest that a longitudinal analysis of facial expression would give reliable results rather than a cross-sectional analysis.
    
    \item Figure ~\ref{fig:f2} shows that a small tilt of the face can be corrected without incurring much error. Using geometric transformation, the face is rotated and aligned, and then the same methodology is followed as in section \ref{algo}.
    
    \item Figure ~\ref{fig:f3} shows that if the tilt of the face is too much (subject looking away from the camera), then the generated composite faces are very different, and hence the error increases for such cases.

    \item The code at \cite{keshariS/FacialAsymmetry} was developed in such a way that it automatically detects tilted faces in specific frames in the input video file and:
    \begin{itemize}
        \item if slightly tilted (up to 5 degrees from vertical): aligns the face before construction of composites
        \item if highly tilted: discards the frame from further computation
    \end{itemize}
    
\end{enumerate}

\FloatBarrier
\FloatBarrier

\section{Discussion}
\label{sec:discussion}
The facial expression of emotions lasts between 0.5 to 4 seconds (Ekman, \cite{porter2008reading}) and sometimes as fast as 33 milliseconds. But they are an important source of information as they reflect concealed emotions. In the blink of an eye, the interviewer may miss perceiving these expressions. Quantitative analysis of such facial emotional expressions is also challenging. Unfortunately, measures that exist to capture this facial expression of emotion, are either too sparse or lack specificity to capture minute changes that happen on the face. We advocate the use of the behaviometry tool- developed on the principles of machine learning to be used as an assistive tool. Behaviometry would help reduce the cognitive demand of the interviewer during the selection process and provide supporting evidence by analyzing facial asymmetry as a function of the expression of different emotions.  The effectiveness of this tool has also been presented by sharing the results of the pilot study that was conducted. Although the individual algorithms are open source, the uniqueness lies in the way they are applied in order to obtain suitable conclusions in the form of complementary evidence about incongruity from the subject being interviewed.

For case I (see Figure ~\ref{fig:f1}) there is not much difference between the emotional expression of the subject in its normal picture in comparison to its L-L and R-R facial composite. But when we compare it with case II (see Figure ~\ref{fig:f2}) and case III (see Figure ~\ref{fig:f3}) we find that the facial composite reflects a distinct change or asymmetry in the L-L and R-R composite in comparison to the normal face. Such changes are often difficult to detect in an ongoing interaction however behaviometry helps to capture these changes and act as an aid to help decision-making. Research suggests that the left side of the face is the social face while the right side of the face reflects true emotion.

It is assumed that during an ongoing interview, an interviewee would try to control the expression of emotion based on certain display rules. At times the interviewees try to impress upon the interviewer by creating a particular image (\cite{roulin2014honest, bourdage2018}). This could be in the form of neutralizing, masking, etc. The use of this behaviometry tool would assist recruiters in the interview selection by identifying facial leakage of the ongoing affective state of an interviewee. However, the advantage of the developed tool may necessarily not be restricted to the personnel selection process, it may well prove to be effective in other situations such as scrutiny, detection of mental health issues, detecting deception, establishing trust, or interacting more humanely.

\subsection{Limitation}
The present study limits its domain to the identification and analysis of facial expressions. However, this method could be extended to the understanding of other nonverbal cues such as vocals, gestures, posture, gait, etc., and also include multi-modal machine learning methods. Secondly, the literature suggests that cultural specificity influences the facial expression of emotion. This was not considered in the present study. A future study addressing this concern may be conducted.

{\small
\bibliographystyle{ieee_fullname}
\bibliography{refs.bib}
}

\end{document}